\documentclass[letterpaper, 10 pt, conference]{ieeeconf}  

\IEEEoverridecommandlockouts                              

\usepackage{times} 
\usepackage{amsmath} 
\usepackage{amssymb}  

\usepackage{graphicx}
\usepackage[caption=false]{subfig}
\usepackage{csquotes}
\usepackage{color}
\usepackage{multirow}
\usepackage{booktabs}
\usepackage[normalem]{ulem} 
\usepackage[ruled,vlined]{algorithm2e}

\newcommand{\R}{\mathbb{R}}

\title{\LARGE \bf
Self-supervised deep visual servoing \\for high precision peg-in-hole insertion}

\author{Rasmus Laurvig Haugaard, Anders Glent Buch, and Thorbj{\o}rn Mosekj{\ae}r Iversen
\thanks{
This work was supported by Innovation Fund Denmark through the project MADE FAST. \newline
All authors are from SDU Robotics, Maersk Mc-Kinney Moller Institute, University of Southern Denmark.\newline
{\tt\small \{rlha,anbu,thmi\}@mmmi.sdu.dk}}
}

\begin{document}
\maketitle
\thispagestyle{plain}
\pagestyle{plain}

\begin{abstract}
Many industrial assembly tasks involve peg-in-hole like insertions with sub-millimeter tolerances which are challenging, even in highly calibrated robot cells. Visual servoing can be employed to increase the robustness towards uncertainties in the system, however, state of the art methods either rely on accurate 3D models for synthetic renderings or manual involvement in acquisition of training data. We present a novel self-supervised visual servoing method for high precision peg-in-hole insertion, which is fully automated and does not rely on synthetic data. We demonstrate its applicability for insertion of electronic components into a printed circuit board with tight tolerances.
We show that peg-in-hole insertion can be drastically sped up by preceding a robust but slow force-based insertion strategy with our proposed visual servoing method, the configuration of which is fully autonomous.


\end{abstract}


\section{Introduction}
\label{sec:intro}

Many industrial insertion tasks require high precision for successful completion. When the insertion tolerances are near or lower than the accumulated uncertainties in the system, naive planning-based insertion becomes unreliable. This can be handled either by decreasing the uncertainties in the system or increasing the tolerances.

For some tasks careful calibration is sufficient to reduce the system uncertainties to within the tolerances. However, since the uncertainties in a system depend on the combined effect of many different factors which are hard to accurately model, there is a practical limit to the accuracy of a system. For insertion tasks with very small tolerances, good calibration is, therefore, often not enough.

A common method for increasing the tolerances of a an insertion is to introduce either active or passive compliance perpendicular to the insertion direction. However, during the search, before there is peg-hole contact, there is no force-feedback perpendicular to the insertion direction, and prior to peg-hole contact, the force-feedback is thus limited to the binary feedback of whether the peg has hit the surface surrounding the hole, or has been, at least partially, inserted.


Binary force feedback can be used for robust insertion as part of an exhaustive search within the region of uncertainty around the estimated insertion point, e.g. using spiral search. While this has been shown to work successfully for PCB assembly~\cite{mathiesen2021towards}, it is a slow technique if the error is significant compared to the tolerances. 
Since the area that must be explored increases quadratically with the magnitude of the error, $|e|$, and the area of successful insertion positions increase quadratically with the tolerance, $\epsilon$, the time for insertion, using an exhaustive search, is proportional to the squared ratio between the error and tolerance,
\begin{equation}
    t \propto \left[ \frac{|e|}{\epsilon}\right]^2.
\end{equation}
Consequently, the insertion time becomes prohibitively large when tolerances are low relative to the system uncertainties.

\begin{figure}[]
    \centering
    \includegraphics[width=\linewidth, trim={0, 0, 224, 224}, clip]{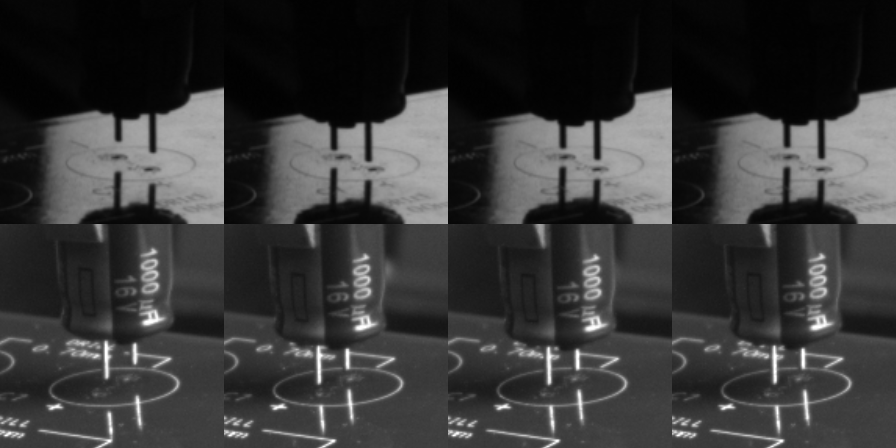}\vspace{2pt}
    \includegraphics[width=\linewidth, trim={0, 0, 224, 224}, clip]{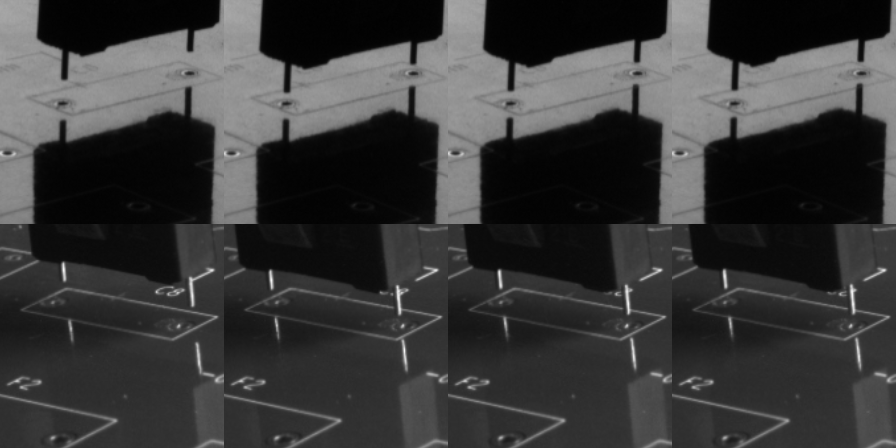}\vspace{2pt}
    \includegraphics[width=\linewidth, trim={0, 0, 224, 224}, clip]{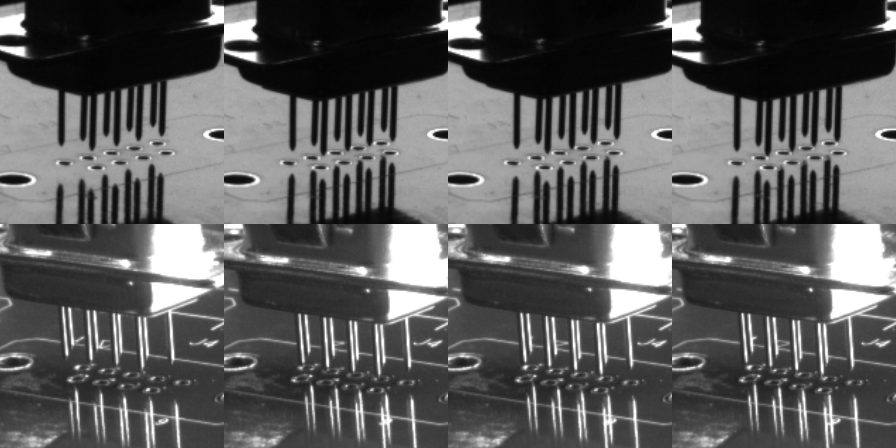}
    \caption{
        Examples of two iterations of visual servoing on three PCB components. 
        From left to right: before visual servoing, after first iteration, after second iteration.
    }
    \label{fig:vs_examples}
\end{figure}

Visual servoing is a technique by which a robot is controlled to reduce the error $e$ based on visual feedback. We argue that visual servoing and exhaustive search should be combined to leverage 
the robustness of slow force-feedback, 
and visual servoing's ability to reduce the error,
to obtain fast and robust insertions.

\begin{figure*}
    \centering
    \includegraphics[width=\linewidth, trim={0, 100, 0, -100}, clip]{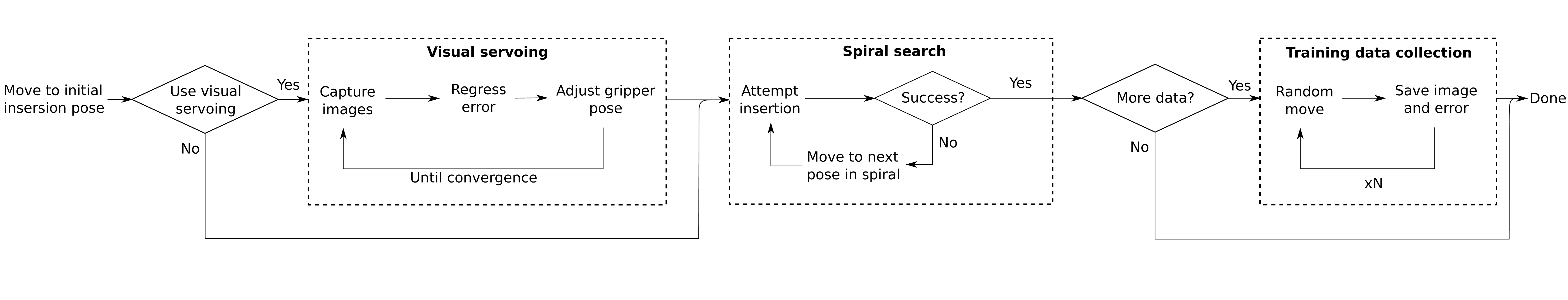}
    \caption{
        Overview of proposed system. 
        For the initial insertions, visual servoing is skipped, and spiral search is used to gather training data autonomously.
        Concurrently, while the system is running, models are trained on the gathered data, and the validation error informs when to enable visual servoing.
        }
    \label{fig:flow}
\end{figure*}

\begin{figure}
    \centering
    \includegraphics[width=\linewidth]{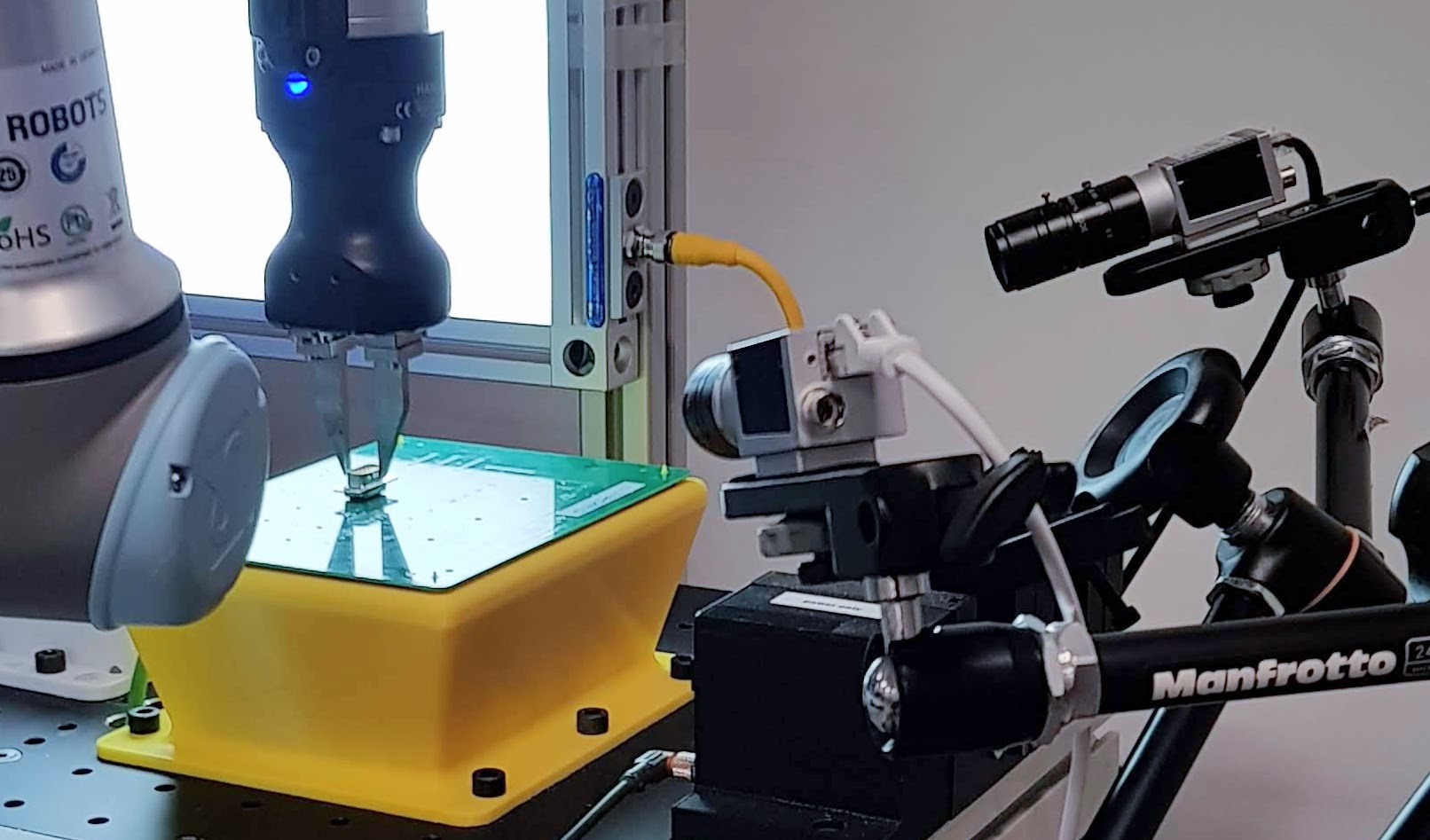}
    \caption{
        The experimental setup with a robot, a TCP-mounted gripper, a PCB fixture, and two cameras.
        The specific light source is designed for another process in the cell and is not necessary for our method.
    }
    \label{fig:system}
\end{figure}

The literature on visual servoing reports that deep learning based methods can be successfully applied to peg-in-hole insertion. However, the methods rely either on representative 3D models for synthetic renderings or manual involvement in the acquisition of annotated training images. The reliance on accurate models or manual configuration limits the practical use and scalability of such methods.

We present a visual servoing method which is fully automated and does not require a model of the object. The proposed method constrains the servoing task to an in-plane 2D alignment of the peg relative to the hole. Our method uses the slow but robust strategy of spiral search to drive self-supervised learning of in-plane alignment by regression of the alignment error in each camera of a multi camera setup.
In our experimental setup, we use two cameras. See Fig.~\ref{fig:system}. 

Our contributions are:
\begin{itemize}
    \item an in-plane deep learning based visual servoing method for  
    high precision peg-in-hole insertion 
    \item and a self-supervised learning scheme based on spiral search which fully automates the configuration and learning process with no reliance on a 3D model.
\end{itemize}

The proposed method is evaluated for the insertion of five different types of electronic components into a PCB. The evaluation shows that our proposed method enables significantly faster peg-in-hole insertion with sub-millimeter precision.

The paper is organized as follows: Section \ref{sec:related} discusses the recent literature related to high precision deep visual servoing. Section \ref{sec:method} then presents the proposed method. This is followed by section \ref{sec:experiments}, which presents the experimental setup, the evaluation procedure, and the results. Finally, \ref{sec:conclusion} presents the conclusion of the work.

\section{Related work}
\label{sec:related}
Visual servoing is defined as a technique by which a robot is controlled by visual feedback~\cite{siciliano2009force}. A control loop determines the movement of a robot's end effector such that an error function, based on visual input, is minimized. The visual input comes from vision sensors which are either mounted on the robot (eye-in-hand) or next to the robot (eye-to-hand). Visual servoing techniques are traditionally categorized depending on the domain in which the error function is defined. Image based visual servoing (IBVS) defines the error function in image space while position based visual servoing (PBVS) defines the error function in eucledian space~\cite{thuilot2002position}.

Most IBVS techniques fall into one of two categories. The first is feature based methods, which defines the error function in terms of the difference between extracted and desired visual features such as keypoints~\cite{weiss1987dynamic}. The second is direct visual servoing (DVS), in which the error function is not defined explicitly on extracted features but instead defined based on the difference between the current image and a target image~\cite{collewet2008visual}.

Recent advances in visual servoing rely on deep learning. While a part of the visual servoing literature focuses on achieving large basins of convergence or high generalizability (e.g. \cite{harish2020dfvs, sadeghi2018sim2real}), the following focuses on literature in which the main objective is the ability to handle tight tolerances for grasping or insertion.

There are several recent methods which use deep learning for DVS. One such method is \cite{felton2022visual} which uses a convolutional auto-encoder to extract a learnt latent space representation of current and target image. The robot is then controlled using a derived interaction matrix, which relates difference in current and target latent vector to a desired change in robot motion. Other works uses siamese networks to extract latent features followed by either a network which regresses the velocity vector of the camera~\cite{felton2021siame} or the relative pose between the current and desired image~\cite{yu2019siamese, tokuda2021convolutional}. The evaluation presented in \cite{yu2019siamese} demonstrates that successful insertion of a tool mounted male sub-D connector (peg) into a female sub-D connector (hole) can be achieved using this scheme. However, unlike our method, they fixate the peg to the robot TCP and only perform servoing with respect to the hole, effectively assuming that the only error is in the hole pose. This approach requires low uncertainty on the peg's in-hand pose which cannot be guaranteed in general for tasks where objects are grasped by an end effector rather than rigidly mounted on the robot. Furthermore, their acquisition of training data requires initial manual guidance of the robot to the ground truth pose, in order to obtain a target image and initialize the acquisition of training images.
Note that the need for a target image is at the core of DVS techniques.
This assumption about a single target image is problematic in cases where within-class variations render the information provided by a single target insufficient.
In case of electrical components for example, a single target image does not provide information about whether to align with respect to the pins or the body of the component, when the pins are slightly bent.
Our method does not assume a single target image and can thus capture such information from multiple targets.

Visual servoing for peg-in-hole insertion has also been done using keypoints extracted with deep learning. Two recent works show that training on synthetic images, combined with heavy domain randomization, can bridge the sim to real gap sufficiently for peg in hole insertions on selected cases. In \cite{haugaard2021fast} two cameras are mounted in an eye-in-hand setup. The center of a hole and the tip of a tool mounted cylindrical peg are regressed using convolutional neural networks (CNN), and from these points the euclidian error is computed and used to control the robot end effector. While the peg in hole task is successfully performed, it is only demonstrated to work on insertion of simple cylindrical objects, matching the geometry of the peg model used in the synthetic data. In \cite{puang2020kovis}, two networks are used: an encoder-decoder network for self-supervised learning of keypoint extraction and a fully connected network for regressing required robot motion from keypoints extracted from an eye-in-hand stereo camera setup. However, the insertion tasks have high tolerances and inherent mechanical compliance to help guide the insertion. Unlike our method, both \cite{haugaard2021fast} and \cite{puang2020kovis} train on synthetic data and thus rely on 3D models and successful sim to real transfer.
While the sim to real transfer has been shown to work on selected cases, 
it depends on the quality of the 3D models and the synthetic renders in general and it is thus not trivial to estimate whether or not it will work for a given task, since the error on even a synthetic validation dataset does not necessarily represent the performance on real data.
Also, when sim to real transfer fails, expert knowledge is required to improve the system. 
In contrast, our training and validation data is drawn from the target distribution, and we are thus able to leverage standard machine learning practices for reliable model evaluation before deployment.

To the best of our knowledge, our work is the first visual servoing method to demonstrate sub-millimeter precision peg-in-hole insertion of non-trivial objects, requiring neither 3D models for synthetic image generation nor manual involvement in the acquisition of training images.

\section{Method}
\label{sec:method}
This section first defines and formalizes the in-plane visual servoing task, 
then presents our method, including 
how we formulate in-plane visual servoing as a well-posed learning problem,
the visual servoing control loop,
and the strategy for self-supervision.

\begin{figure}
    \centering
    \includegraphics[width=1\linewidth, trim={170, 240, 340, 20}, clip]{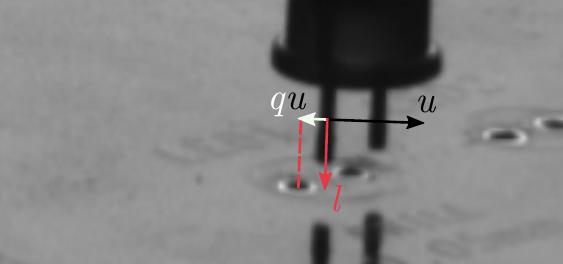}
    \caption{
        The insertion direction, $l$, is shown with a red arrow, 
        the error direction, $u$, is shown in black, 
        and $qu$ is shown in white, 
        all projected into the image for visualization.
    }
    \label{fig:vectors}
\end{figure}

\subsection{In-plane visual servoing}
\label{subsec:formalize}
We define in-plane visual servoing as the control of a robot where the movement of the end effector is constrained to a constant rotation and a displacement confined to a plane, that is $\frac{d\phi}{dt}=0$ 
and 
$\frac{dp}{dt}\cdot l=0$
, where $\phi\in \text{SO(3)}$ is the rotation, 
$p\in \mathbb{R}^3$ is the TCP position, and 
$l\in\mathbb{R}^3$ is the insertion direction and thus a normal vector to the alignment plane.



Constraining the visual servoing task to positional, 2-DOF, alignment is a simplification compared to full 5-DOF alignment, followed by insertion along the last DOF. 
However, small errors in the 3-DOF rotation can be seen as a reduction of the insertion tolerances,
and orientation constraints from mechanical feeders and grippers help to reduce these errors.
We thus argue that the simplification is valid in many cases, and hypothesize that the simplification leads to faster and more robust visual servoing.

\subsection{Proposed visual servoing method}
\label{subsec:vsmethod}
The prerequisite for our method is a calibrated setup, like the one seen in Fig.~\ref{fig:system}, with a robot, a gripper attached to the robot TCP and two or more cameras. The cameras in our experimental setup are attached to the table, but can be attached in-hand or to a separate robot for increased flexibility.
Our method relies on system calibration to extract crops and to relate the insertion direction to the images, but the visual feedback loop ensures that the method is robust towards calibration uncertainties.

The reason for choosing at least two cameras is to create a well posed regression problem.
The in-plane alignment error could theoretically be regressed from a single image by relative depth estimation from perspective effects.
However, since the objects are small compared to the distance between camera and insertion, the camera projection is close to orthographic in the region of interest.
It could also be argued that the in-plane alignment of the peg can be learned from the peg's projection onto the image plane relative to the projection of the background. However, this would introduce strict requirements for the distance between hole plane and peg to be constant, which in turn would reduce the robustness with respect to uncertainties, including but not limited to grasping uncertainties, part variations and time-dependent calibration errors.
Consequently, we argue that refraining from regressing the depth alignment and instead obtain the in-plane alignment from triangulation from two or more cameras, as in \cite{haugaard2021fast}, is a significantly better posed problem.
We hypothesize that a well-posed servoing problem is key to learning a mapping that is robust to uncertainties and generalizes well to new instances.

Given an insertion direction, $l\in\R^3$, the position of a camera, $c\in\R^3$, and the approximate position of the insertion, $x\in\R^3$,
we define the view vector, $v=x-c$, and the camera specific error direction, $u=\frac{l\times v}{|l\times v|}$.
Given an in-plane alignment error, $e\in\R^3$, $e \cdot l = 0$, the scalar error along the error direction is $q = e \cdot u$.
See Fig~\ref{fig:vectors}.
Under the assumption of orthographic projection in the region of interest, the error can be reconstructed, like in \cite{haugaard2021fast}, as the least squares solution to the set of linear equations established by scalar errors from multiple views,
\begin{equation}
    \begin{bmatrix}
        u_{1x} & u_{1y} & u_{1z} \\
        u_{2x} & u_{2y} & u_{2z} \\
        & \vdots
    \end{bmatrix}
    \begin{bmatrix}
    e_x \\ e_y \\ e_z
    \end{bmatrix}
    =
    \begin{bmatrix}
    q_1 \\ q_2 \\ \vdots
    \end{bmatrix}.
\end{equation}

We propose to learn a mapping, $f_\theta: \R^{r\times r} \mapsto \mathbb{R}$, from a camera image, $I$, with resolution $r$ to the scalar error, $q$, however, to make the dynamic range of the output independent of object scale and image resolution, we learn the mapping to $y=qf(rz)^{-1}$ instead, where $f$ is the focal length in pixels, $z$ is the approximate depth of the insertion in the camera, and $y\in\R$ is the normalized scalar error in the image.
Specifically, we aim to learn the parameters, $\theta$ of a CNN, $f_\theta$, that minimizes the mean squared error loss
\begin{equation}
    L = \frac{1}{N}\sum_i^N (y_i - f_\theta(I_i))^2.
\end{equation}

During inference, $q$ can be established from $y$. The full visual servoing control loop is shown in Algorithm~\ref{alg:visual-servo}.

\begin{algorithm}
\DontPrintSemicolon
\SetAlgoLined
    \KwIn{
        Insertion direction, $l$.
        Approximate hole position, $h$.
        Approximate camera positions, $c_j$ and focal lengths, $f_j$, of $n_\text{cams}$ cameras.
        Number of iterations, $n_\text{iters}$.
    }
    $p \gets$ current TCP position\;
    \For{$i = 1, ..., n_\text{iters}$}{
        \For{$j = 1, ..., n_\text{cams}$}{
            $I_j \gets$ capture image from camera $j$\;
            $y_j \gets f_\theta(I_j)$\tcp*{est. norm. error}
            
            $v_j \gets h - c_j$\tcp*{view vector}
            $u_j \gets \dfrac{l \times v_j}{||l \times v_j||}$\tcp*{error direction}
            $q_j  \gets y_j r z f^{-1} $\tcp*{error magnitude}
        }
        $(U, q) \gets \left( 
            \begin{bmatrix}
                u_{1x} & u_{1y} & u_{1z} \\
                u_{2x} & u_{2y} & u_{2z} \\
                & \vdots
            \end{bmatrix}
            ,
            \begin{bmatrix}
                q_1 \\ q_2 \\ \vdots
            \end{bmatrix}
            \right)
        $\;
        $\hat{e} \gets$ solve $U\hat{e}=q$ by least squares\;
        $p \gets p + \hat{e}$\;
        move TCP to $p$\;
    }
    \caption{In-plane visual servoing}
\label{alg:visual-servo}
\end{algorithm}

\begin{figure}
    \centering
    \includegraphics[width=0.5\linewidth, trim={0, 0, 0, 0}, clip]{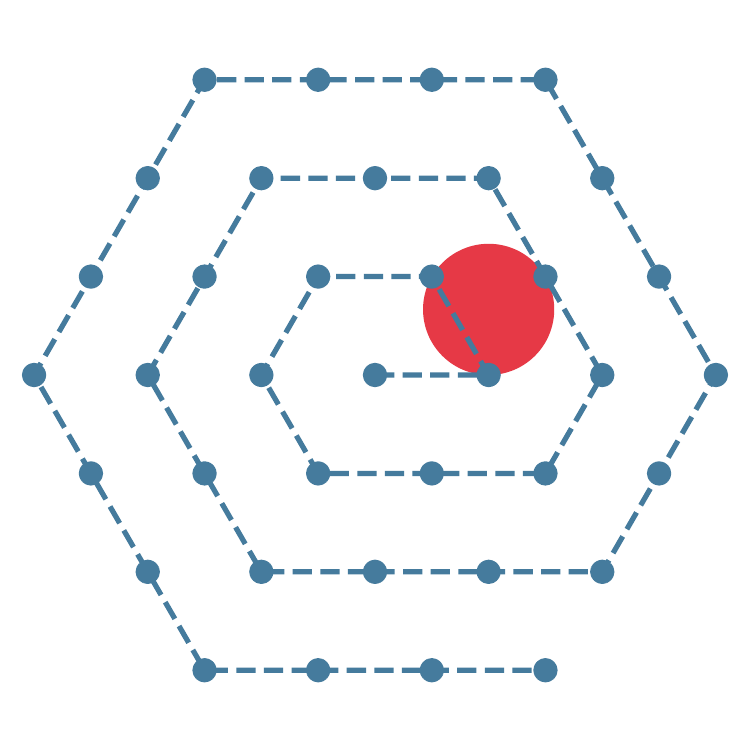}
    \caption{
        Our in-plane spiral search pattern in blue.
        The largest enclosed circle is drawn in red. 
        The circle diameter represents the minimum tolerance for which the pattern is guaranteed to succeed.
    }
    \label{fig:honey_comb_search}
\end{figure}

\subsection{Autonomous configuration and continuous learning}
\label{sec:auto-config}

The visual servoing network presented in Sec.~\ref{subsec:vsmethod} could in principle be trained using manually annotated images and/or synthetic images. However, the strength of our method lies in having autonomous collection of training data from the target distribution. This allows the configuration of the visual servoing method for previously unseen objects to be reliable and fully autonomous.
See Fig.~\ref{fig:flow}.

Peg-in-hole insertion requires poses of peg and hole, and depending on the system, this is approximately known based on calibration, demonstration, pose estimation, vibrational feeders, and/or by other means. Assuming that the tolerance for insertion is lower than the uncertainty on the relative hole to peg pose, a search strategy is needed to ensure successful insertion, and when such a method is in place, like spiral search, the insertion is guaranteed given enough time.

When an insertion has been successfully completed,
we know that the end TCP position led to a successful insertion, and thus we can assume that to be the correct in-plane position.
Since most industrial robots have a low \textit{relative} positional uncertainty, it is possible to acquire annotated training data by capturing images with \textit{known} in-plane errors. From such a dataset it is straight forward to compute the ground truth scalar error, $y$, in each image.

We can relax the previous assumption that the in-plane position at \textit{a} correct insertion position is \textit{the} correct, fully centered position.
The mean squared error loss assumes that $p(y|I)$ is normally distributed, and leads to regressing the mean of said distribution. Even though the insertions in the training data are not fully centered, the vision model should still learn to regress the center, as long as the 
assumption that $p(y|I)$ is normally distributed around the true $y$ is good.

The autonomously obtained dataset is split into a training and a validation set, and the model is trained with early stopping, monitoring the loss on the validation set to inform when to stop to avoid overfitting.
We split the data such that all data points from an insertion either goes to the training or the validation set to enable early stopping that specifically monitors the model's ability to generalize to new insertions, in contrast to sampling the validation set randomly from all data points.
The performance on the validation set can be used to decide whether the model should be deployed on the system, or if more data is required to obtain a good model.

During deployment, the performance of the model can also be monitored to detect a distributional shift which could be caused by changes in the objects, e.g. due to supplier-change, or by changes to other parts of the system, like a new background, lighting, time-dependent calibration errors, etc.
The system would then be able to collect more data and autonomously obtain robustness to the variations that occur in the system.

\section{Experiments}
\label{sec:experiments}

\begin{figure}
    \centering
    \includegraphics[width=\linewidth, trim={0, 60, 0, 0}, clip]{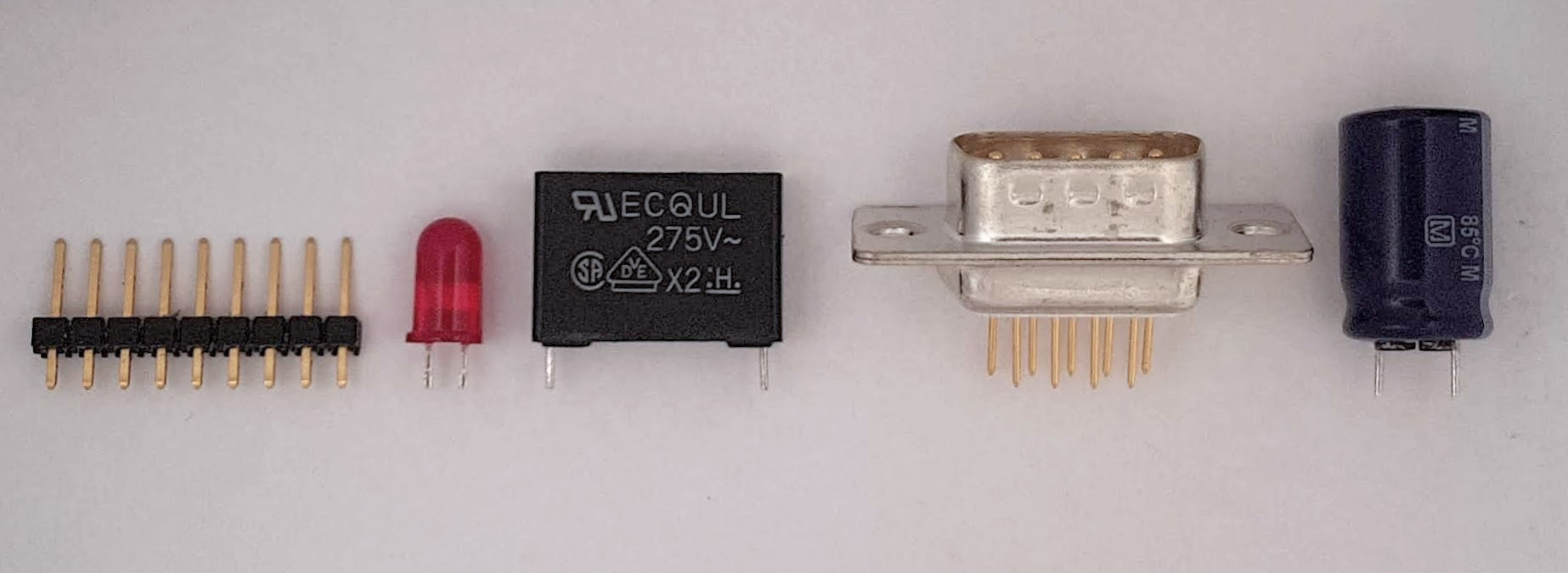}
    \caption{
        The five electronic components used in the experiments.
        We refer to them, from left to right, as PH, LED, C1, DSUB and C2.
    }
    \label{fig:components}
\end{figure}

We evaluate our visual servoing method on insertion of electronic components into a PCB.
The experimental setup is shown in Fig.~\ref{fig:system}, and we choose five different types of components, shown in Fig.~\ref{fig:components}, covering variance in size, number of pins and visual appearance.
The insertion tasks have sub-millimeter tolerances, lower than the accumulated errors in system calibration, grasp uncertainties, and PCB tolerances, making it a relevant case for our method.
The setup has mechanical fixtures and linear vibrational feeders, from which the robot can grasp the components.

As introduced in Sec.~\ref{sec:intro}, spiral search provides a robust but slow way to deal with the fact that the uncertainties are larger than the insertion tolerances.
A common spiral search implementation lets the peg slide on the surface surrounding the hole while applying a force in the insertion direction, registering if the peg dips into the surface, indicating an at least partial insertion.
In our case however, a such approach could bend the pins and damage the PCB surface.
Instead, our spiral search implementation attempts insertions in a spiral-like pattern at the intersections of an isometric grid, as shown in Fig.~\ref{fig:honey_comb_search}.
This, like the common archimedean spiral search, provides guarantees for the diameter of the largest enclosed circle, the minimum allowed tolerance, and is more efficient than a regular grid.

As discussed in Sec.~\ref{sec:auto-config}, we use the robust but slow search to obtain visual servoing training data autonomously.
See Fig.~\ref{fig:flow}.
Specifically, we perform ten insertions for each component type and capture images at 100 sampled TCP positions per insertion.
The TCP positions are sampled around the successful in-plane position. We sample the direction of the error along the plane uniformly, the magnitude along the direction uniformly from 0 to 1~mm, and an offset perpendicular to the plane, also uniformly between 0 and 1~mm.

For each component type, a vision model is trained to predict $y$, as discussed in Sec.~\ref{subsec:vsmethod}.
Any capable vision architecture can be used.
In our experiments, we use a ResNet50~\cite{resnet} pretrained on ImageNet~\cite{imagenet} classification, replacing the last fully connected layer with a two-layer MLP for regressing $y$. 
The MLP has 128 neurons and ReLU activations at the pre-ultimate layer.
The models are trained with the Adam~\cite{adam} optimizer and a learning rate of $10^{-4}$.
We use the data from eight of the insertions for training and two for validation to enable early stopping that specifically monitors the model's ability to generalize to new insertions.

The total configuration time per component was approximately 25 minutes, including 10 minutes for data collection and 15 minutes for training on an RTX2080 GPU.

\begin{table}
    \centering
    \small
    \caption{
        Mean insertion times in seconds \\with and without visual servoing.
    }
    \hskip.5cm\begin{tabular}{l|rrrrr|r}
        & PH & LED & C1 & DSUB & C2 & Avg   \\
    \toprule
    vs & 1.7   & 1.8    & 2.0   & 1.7   & 2.3 & 1.9  \\
    no vs & 11.8  & 47.5   & 42.7  & 11.3  & 47.8 & 32.2  \\
    \bottomrule
    \end{tabular}
    \label{tab:main_results}
\end{table}

\begin{figure}
    \centering
    \includegraphics[width=1\linewidth, trim={0, 0, -20, -10}, clip]{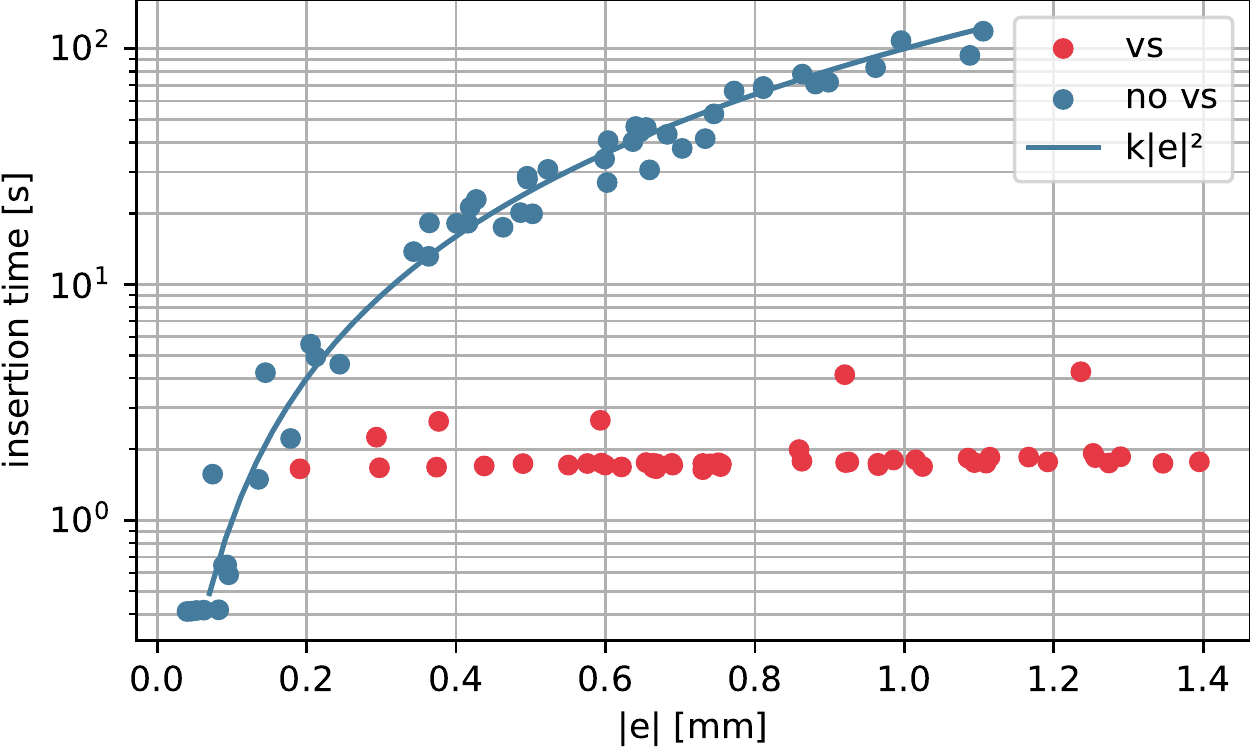}
    \caption{
        Insertion times as a function of retrospective error.
        Note the logarithmic scale on the time axis.
        $k$ is a constant.
    }
    \label{fig:error_time_scatter}
\end{figure}

To evaluate our method's robustness and speed with respect to uncertainties, we sample errors uniformly from an in-plane disc with a radius of 1~mm, in addition to the inherent system uncertainties.
We then attempt spiral search insertion with and without preceeding visual servoing. 
For both approaches we attempt ten insertions per component, amounting to 100 insertions in total. We use three iterations of visual servoing and a spiral search tolerance of 0.1~mm for all insertions.

The insertion times are presented in Table~\ref{tab:main_results}, where insertion time refers to the duration from when the TCP is immediately above the hole in the sampled error position with the grasped component, to when a successful insertion is determined by spiral search.
We perform the spiral search with up to 1~mm error, which is the size of the error we sample during the experiments, however, since the error is added to the accumulated errors inherent in the system, 2 of the 50 spiral search insertions without visual servoing fails.
All the insertions with visual servoing succeeds, while being significantly faster across all component types and more than 15 times faster on average compared to insertion without visual servoing.

Because the sampled errors are added to the system uncertainties, we do not know the actual error a priori, however, like we obtain the dataset, we can use the position at successful insertion to estimate the initial error, retrospectively.
The relationship between the retrospective initial error and the insertion time is visualized in Fig.~\ref{fig:error_time_scatter} for all 100 insertions.
Pure spiral search is fast, when the error is very small, but increases quadratically with the error.
In contrast, the insertion time with visual servoing is approximately constant.

Note that the retrospective initial errors are estimates and are only exact up to the actual insertion tolerances which are different between the five component types, and largest for PH and DSUB, which also shows in Table~\ref{tab:main_results}.
Insertion based solely on spiral search will tend to find the closest successful insertion, explaining the generally lower retrospective errors without visual servoing.

The critical error, where visual servoing outperforms pure spiral search, depends on the tolerances of a given task. Fig.~\ref{fig:error_time_scatter} indicates that visual servoing leads to a reduction in insertion time when the accumulated system uncertainty is more than approximately 0.2~mm.

41 of 50 insertions with visual servoing inserts directly at the center point of the spiral search, compared to 5 of 50 insertions without visual servoing. 
The average retrospective error after visual servoing is 0.03~mm.

Of the average insertion time of 1.9~s with visual servoing, 1.3~s is spent on visual servoing itself and the remaining 0.6~s is spent on the spiral search.
Of the 1.3~s spent on visual servoing, 0.5~s is spent on capturing images, 0.4~s is spent on forward passes on the vision model, and 0.4~s is spent on physically moving the robot.
We use rather long exposure times, capture the images sequentially, and run the model inference on a laptop CPU.
Adding a bright light source, capturing the images in parallel and running inference on a GPU could thus reduce the time spent on image acquisition and inference.

Also, continuous visual servoing, as proposed in \cite{haugaard2021fast} would be able to further reduce the impact of image acquisition time, because images are captured in parallel with robot motion, and reduce the time for robot motion, since the robot does not need to come to a stop for image acquisition.
Note however, that the increase in speed comes with added implementation complexity in terms of image timestamp synchronization, and the visual servoing path's dependence on acquisition and inference speed.



\section{Conclusion}
\label{sec:conclusion}

This paper presented a novel self-supervised deep visual servoing method for high precision peg-in-hole insertion. 
The method is fully automated and does not rely on the availability of 3D models.
This is achieved by constraining the visual servoing task to in-plane alignment and training a convolutional neural network to regress scalar alignment errors in image space in a dual camera setup.
Annotated data is gathered autonomously using a robust but slow force-based search method.
The method has been evaluated on insertion of PCB components. 
The evaluation showed that preceding a robust but slow search strategy with our proposed method reduced the average insertion time by an order of magnitude.

{\small
\bibliographystyle{IEEEtran}
\bibliography{references}
}

\end{document}